\pdfoutput=1

\documentclass[11pt]{article}

\usepackage[final]{acl}

\usepackage{times}
\usepackage{latexsym}

\usepackage[T1]{fontenc}

\usepackage[utf8]{inputenc}

\usepackage{microtype}

\usepackage{inconsolata}

\usepackage{graphicx}
\usepackage{graphics}
\usepackage{amsmath}
\usepackage{subcaption}
\usepackage{rotating}
\usepackage{titlesec}
\newcommand{\secname}[1]{\noindent{\textbf{#1}}\hspace{0.5em}}
\titlespacing*{\section}{0pt}{1em}{0.3em}
\titlespacing*{\subsection}{0pt}{1em}{0.3em}

\title{Is Semantic Chunking Worth the Computational Cost?}

\author{Renyi Qu \\ Vectara, Inc. \\ \texttt{renyi@vectara.com} \\\And
        Forrest Bao \\  Vectara, Inc. \\ \texttt{forrest@vectara.com} \\\And
        Ruixuan Tu \\ University of Wisconsin--Madison\\ \texttt{turx2003@gmail.com}}

\begin{document}
\maketitle
\begin{abstract}
Recent advances in Retrieval-Augmented Generation (RAG) systems have popularized \textbf{semantic chunking}, which aims to improve retrieval performance by dividing documents into semantically coherent segments. Despite its growing adoption, the actual benefits over simpler \textbf{fixed-size chunking}, where documents are split into consecutive, fixed-size segments, remain unclear. 
This study systematically evaluates the effectiveness of semantic chunking using three common retrieval-related tasks: document retrieval, evidence retrieval, and retrieval-based answer generation. 
The results show that the computational costs associated with semantic chunking are not justified by consistent performance gains. These findings challenge the previous assumptions about semantic chunking and highlight the need for more efficient chunking strategies in RAG systems.
\end{abstract}

\section{Introduction}
In Retrieval-Augmented Generation (RAG) systems, cutting documents into smaller units called ``chunks'' has a crucial effect on the quality of both retrieval and generation tasks \cite{densexretrieval, rahul, llmdistracted, chainofnote}. By retrieving the most relevant chunks for a given query and feeding them into a generative language model, these systems aim to produce accurate and contextually appropriate answers. However, the effectiveness of chunking strategies remains a significant challenge in optimizing retrieval quality and computational efficiency \cite{rag, rag2}.

Known as \textbf{fixed-size chunking}, the traditional way to chunk is to cut documents into chunks of a fixed length such as 200 tokens~\cite{rag3}. While computationally simple, this approach can fragment semantically related content across multiple chunks, leading to suboptimal retrieval performance. Recently, there has been a surge of interest in \textbf{semantic chunking}, where documents are segmented based on semantic similarity, with some industry applications suggesting promising improvements in performance \cite{langchain, llamaindex, dstar}. 
However, there is no systematic evidence that semantic chunking yields a  performance gain in downstream tasks, and if there is, the gain is significant enough to justify the computational overhead than fixed-size chunking. 

Such a systematic evaluation is not trivial due to the lack of data that can be directly used to compare  chunking strategies. 
Therefore, we design an indirect evaluation using three proxy tasks: (1) document retrieval, measuring the ability to identify relevant documents; (2) evidence retrieval, measuring the ability to locate ground-truth evidence; and (3) answer generation, testing the quality of answers produced by a generative model using retrieved chunks. Our findings challenge prevailing assumptions about the benefits of semantic chunking, suggesting that its advantages are highly task-dependent and often insufficient to justify the added computational costs. This study lays the groundwork for future exploration of more efficient and adaptive chunking strategies in RAG systems.

\begin{figure*}[ht]
  \includegraphics[width=\linewidth]{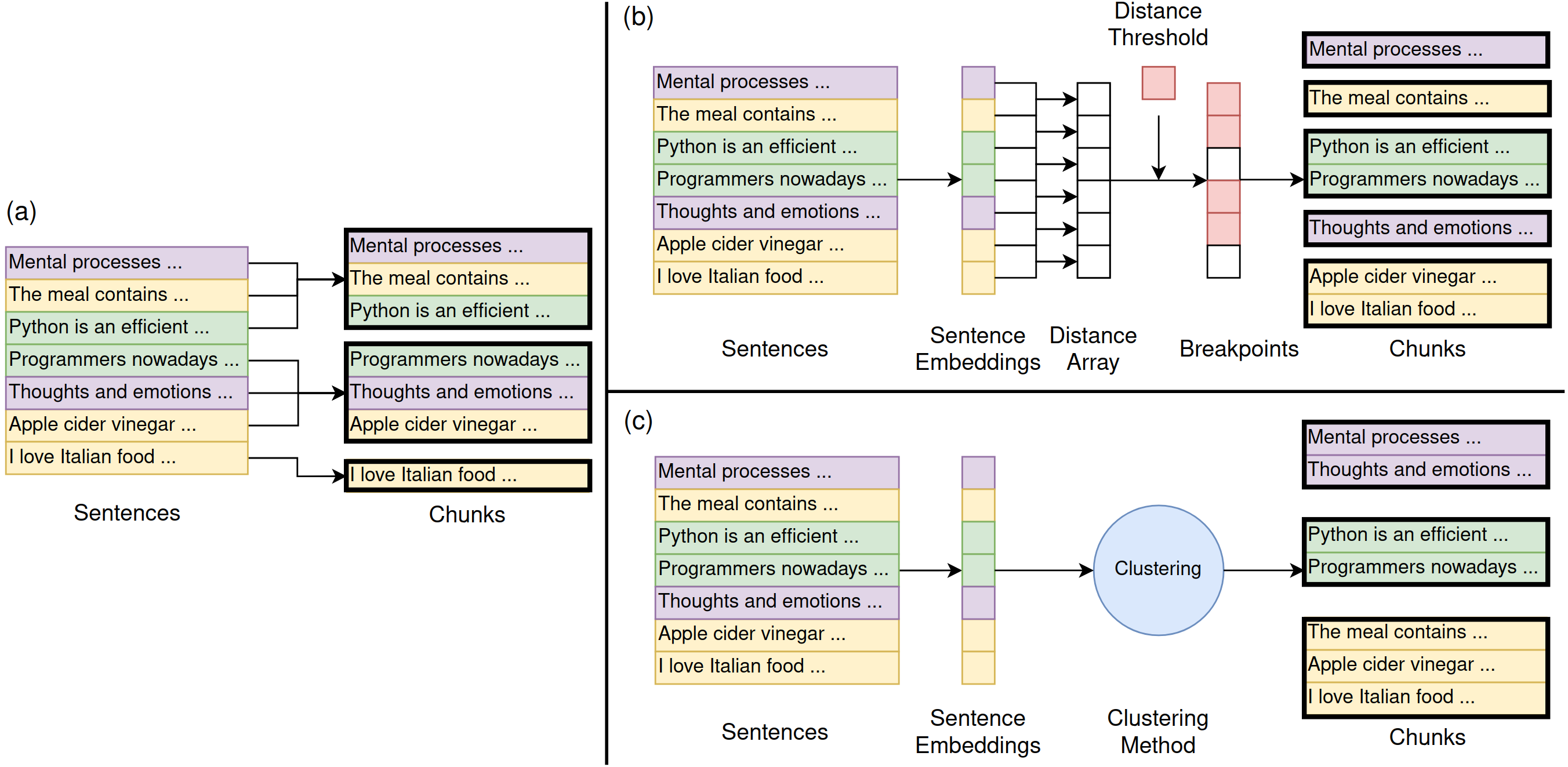}
  \caption{Illustration of the three chunkers tested in this study. Colored segments represent different topics within the sample document: Purple for psychology, Green for programming, and Yellow for food. Red blocks mark chunk breakpoints. (a) Fixed-size Chunker splits the document into consecutive, uniform chunks without considering semantic content. (b) Breakpoint-based Semantic Chunker segments the text by detecting semantic distance thresholds between consecutive sentences to maintain coherence. (c) Clustering-based Semantic Chunker groups semantically similar sentences, potentially combining non-consecutive text to form topic-based chunks.}
\end{figure*}

In general, our contributions are:
\begin{itemize}
\item We present a novel, large-scale evaluation framework comparing semantic and fixed-size chunking across diverse tasks.
\item We demonstrate that while semantic chunking shows some benefits in certain scenarios, these are inconsistent and often insufficient to justify the computational cost.
\end{itemize}

\section{Chunking Strategies}
In this paper, a document is first split into sentences which are then grouped into chunks. We evaluate three chunking strategies, hereafter referred to as ``chunkers.'' 

\secname{Fixed-size Chunker}This is our baseline chunker that splits a document sequentially into fixed-size chunks, based on a predefined or user-specified number of sentences per chunk. 

Although this approach is simple and computationally efficient, it may separate contextually related sentences, leading to potential degradation in retrieval quality \cite{rag, rag2, rag3}. To alleviate this, we use overlapping sentences between consecutive chunks, a common practice to maintain some degree of contextual continuity.
\vspace{0.1em}


\secname{Breakpoint-based Semantic Chunker}A break- point-based chunker scans over the sequence of sentences and decide where to insert a breakpoint to separate sentences before and after it into two chunks. 
A breakpoint is inserted if the semantic distance between two consecutive sentences exceeds a thredshold, meaning a significant topic change. 

We tested four relative thresholds for determining breakpoints, as proposed by \citet{langchain}.
Additionally, we tested two absolute thresholds, which use predetermined values to determine chunk boundaries, reducing computational overhead.

However, the breakpoint-based chunkers make decisions using only two sentences each time. This strategy maybe locally greedy. To chunk with more information at a bigger scope, we propose a new type of semantic chunkers next.
\vspace{0.1em}

\secname{Clustering-based Semantic Chunker}
This type of chunkers leverage clustering algorithms to group sentences together semantically, capturing global relationships and allowing for non-sequential sentence groupings. 
However, it risks losing losing contextual information hidden in the proximity of sentences. To mitigate this, we defined a new distance measure that combines positional and semantic distances. Specifically, we calculate a weighted sum between the positional distance (i.e., the sentence index difference) and the cosine distance between two sentence $\textbf{x}_a$ and $\textbf{x}_b$:\vspace{-1em}

{\small\begin{align}
\scriptstyle
&d(\textbf{x}_a,\textbf{x}_b)=\lambda d_{\text{pos}}(\textbf{x}_a,\textbf{x}_b)+(1-\lambda) d_{\text{cos}}(\textbf{x}_a,\textbf{x}_b)\label{eq:1} \\
&d_{\text{pos}}(\textbf{x}_a,\textbf{x}_b)=\frac{|a-b|}{n} \\
&d_{\text{cos}}(\textbf{x}_a,\textbf{x}_b)=1-\max(\cos(\texttt{emb}(\textbf{x}_a), \texttt{emb}(\textbf{x}_b)), 0)\label{eq:3}
\end{align}}

where $n$ is the total number of sentences in the document, $\texttt{emb}(\cdot)$ is the embedding function, and $\lambda$ is a hyperparameter. When $\lambda=0$, the chunker operates purely based on semantic similarity; when $\lambda=1$, it mirrors the Fixed-size Chunker. In Eq.~(\ref{eq:3}), a cosine similarity of 0 indicates orthogonal (unrelated) sentence embeddings, while negative cosine similarity values are treated as 0, as they do not aid in retrieval or generation.

Without losing generality, we employed single-linkage agglomerative clustering and DBSCAN \cite{dbscan} as representatives of clustering algorithms. Further details on these methods an their adjustments during experimentation are provided in Appendix \ref{sec:clustering}.

\section{Experiments}
In the absence of ground-truth chunk data, we designed three experiments to indirectly assess the quality of each chunker: document retrieval, evidence retrieval, and answer generation. Different datasets and evaluation metrics were used for each experiment to align with the specific task requirements. All documents were first split into sentences using SpaCy's \texttt{en$\_$core$\_$web$\_$sm} model \cite{spacy} before being embedded and chunked. We tested three embedding models selected to represent a range of performances based on their rankings on the MTEB Leaderboard \cite{mteb}. See Appendix \ref{sec:embedders} for details. 

\subsection{Document Retrieval}\label{sec:document_retrieval}
This experiment assessed the effectiveness of chunkers in retrieving relevant documents for a given query. We used 10 datasets, shown in Tables \ref{tab:doc_retrieval} and \ref{tab:datasets_doc_ret}. Most documents on the BEIR benchmark \cite{beir} are too short for chunking to be effective. To address this, we synthesized longer documents by stitching short documents from six datasets where documents are too short (see Appendix \ref{sec:data_synthesis} for details). We randomly sampled 100 queries from each dataset and retrieved the top $k$ chunks, where $k \in [1, 3, 5, 10]$. Each retrieved chunk was mapped to its source document, and the retrieved documents were evaluated by comparing them to a set of relevant documents for each query.

\subsection{Evidence Retrieval}\label{sec:evidence_retrieval}
Here we evaluate chunkers at a finer granularity than the previous experiment by measuring their abilities to locate evidence sentences. We selected additional datasets from RAGBench \cite{ragbench}, shown in Tables \ref{tab:evi_retrieval} and \ref{tab:datasets_evi_ret}, because few datasets contain long documents with ground-truth evidence sentences. We measured the number of ground-truth evidence sentences present in the retrieved top-k chunks.

\subsection{Answer Generation} This experiment measured how chunkers impacted the quality of LLM-generated answers. We used \texttt{gpt-4o-mini} as the generative model. The top-5 retrieved chunks were used as input for the LLM, and generated answers were compared to ground-truth responses using semantic similarity measures. We reused the datasets from Section \ref{sec:evidence_retrieval}, as they included long documents, evidence, and reference answers.

\section{Results}\label{sec:results}
\subsection{Measuring and reporting performances}

As mentioned earlier, we used three proxy tasks the study chunking. 
We cannot directly assess the quality of retrieval at the chunk level due to the lack of ground-truth at the chunk level.
Instead, each retrieved chunk is mapped back to either the source document or the included evidence sentences. 

Since the number of relevant documents or evidence sentences is not fixed (unlike the $k$ value for retrieved chunks), traditional metrics such as Recall@k and NDCG@k are not suitable. F1 provides a balanced measure that accounts for both precision and recall under these circumstances. Therefore, we use \textbf{F1@5} as the metric.  For further details, see Appendix \ref{sec:metrics}.

For each dataset, results are reported based on the best hyperparameter configuration for each chunker, determined by the average F1 score across all $k$ values. All results to be reported below are obtained using \resizebox{122pt}{!}{\texttt{dunzhang/stella$\_$en$\_$1.5B$\_$v5}} as the embedder for being the best among those tested.

In the following subsections, \textbf{Bold} values indicate the best performance on the respective dataset. The results for Answer Generation closely matched those of Evidence Retrieval and are discussed in Appendix \ref{sec:generation}. Additional analysis of hyperparameters is provided in Appendix \ref{sec:hyperparam}. Inspection of the outputs of different chunkers is provided in Appendix \ref{sec:inspect}.

\subsection{Document Retrieval}
Table~\ref{tab:doc_retrieval} shows varied chunker performance, with Fixed-size Chunker excelling on non-stitched datasets and Semantic Chunkers performing better on stitched datasets.

\begin{table}[h]
\centering
\resizebox{\columnwidth}{!}{\begin{tabular}{l|r|r|r}
Dataset & Fixed-size & Breakpoint & Clustering \\
\hline
Miracl*       & 69.45 & \textbf{81.89} & 67.35 \\
NQ*           & 43.79 & \textbf{63.93} & 41.01 \\
Scidocs*      & 16.82 & 17.60 & \textbf{19.87} \\
Scifact*      & 35.27 & \textbf{36.27} & 35.70 \\
BioASQ*       & 61.86 & 61.87 & \textbf{62.49} \\
NFCorpus*     & 21.36 & 21.07 & \textbf{22.12} \\
\hline HotpotQA      & \textbf{90.59} & 87.37 & 84.79 \\
MSMARCO       & \textbf{93.58} & 92.23 & 93.18 \\
ConditionalQA & \textbf{68.11} & 64.44 & 65.94 \\
Qasper        & \textbf{90.99} & 89.27 & 90.77 \\
\end{tabular}}
\caption{F1@5 for Document Retrieval ($\%$). Datasets marked with * are stitched. Rows are sorted by the average number of sentences per document (before stitching) in ascending order for easier comparison.}
\label{tab:doc_retrieval}
\end{table}

As described in Appendix \ref{sec:data_synthesis}, stitched documents, averaging 100 sentences, were formed by combining short documents (fewer than 10 sentences) from datasets like Miracl and NQ, leading to high topic diversity. In such cases, Breakpoint-based Semantic Chunker outperformed others by better preserving topic integrity, splitting sentences based on semantic dissimilarity to form chunks similar to the original documents. In contrast, Fixed-size and Clustering-based Chunkers often mixed sentences from different documents, increasing noise and lowering retrieval quality.

As document length increased, fewer documents were stitched together, reducing topic diversity. This diminished the advantage of Breakpoint-based Semantic Chunker, while Clustering-based Semantic Chunker improved. The gap between semantic and fixed-size chunkers narrowed, with Fixed-size Chunker benefiting from higher topic integrity.

These results suggest that in real life, the topics in 
a document may not be as diverse as in our artificially noisy, stitched data, 
and hence semantic chunkers may not have an edge over fixed-size chunker there. 

\subsection{Evidence Retrieval}
As shown in Table \ref{tab:evi_retrieval}, Fixed-size Chunker performed best on 3 out of 5 datasets, indicating a slight edge in capturing core evidence sentences. However, the performance differences between the Fixed-size Chunker and the two semantic chunkers were minimal, suggesting no clear advantage for any specific chunking strategy. See Appendix \ref{sec:hyperparam} for more details.

Further inspection revealed that despite variations in chunking methods, the top-k retrieved chunks frequently contained the same evidence sentences, explaining the minimal performance differences. This suggests that adding semantic information did not significantly enhance performance, as the benefits of semantic grouping were often redundant when core evidence was already captured by sentence positions. These findings indicate that the performance of the chunkers largely depends on how effectively the embedding models capture the semantic richness of individual sentences, rather than the chunking strategy itself.

\begin{table}[h]
\centering
\resizebox{\columnwidth}{!}{\begin{tabular}{l|r|r|r}
Dataset & Fixed-size & Breakpoint & Clustering \\
\hline
ExpertQA      & \textbf{47.11} & 47.08 & 46.87 \\
DelucionQA    & 43.05 & 43.24 & \textbf{43.36} \\
TechQA        & \textbf{28.98} & 28.49 & 27.96 \\
ConditionalQA & 18.23 & \textbf{19.83} & 19.14 \\ 
Qasper        & \textbf{8.66} & 8.16 & 8.50 \\
\end{tabular}}
\caption{F1@5 for Evidence Retrieval ($\%$). Rows are sorted by the average number of sentences per document in ascending order for easier comparison.}
\label{tab:evi_retrieval}
\end{table}\vspace{-1em}

\subsection{Results for Answer Generation}\label{sec:generation}
As shown in Tables \ref{tab:bertscore}, Semantic Chunkers performed slightly better than Fixed-size Chunker based on BERTScore, but the differences are minimal, making it difficult to draw any definitive conclusions.
\begin{table}[h]
\centering
\resizebox{\columnwidth}{!}{\begin{tabular}{l|r|r|r}
Dataset & Fixed-size & Breakpoint & Clustering \\
\hline
ExpertQA      & 0.65 & 0.65 & \textbf{0.65} \\
DelucionQA    & 0.76 & \textbf{0.76} & 0.76 \\
TechQA        & 0.68 & 0.68 & \textbf{0.68} \\
ConditionalQA & 0.42 & \textbf{0.43} & 0.43 \\
Qasper        & 0.49 & 0.49 & \textbf{0.50} \\
\end{tabular}}
\caption{BERTScore for Answer Generation.}
\label{tab:bertscore}
\end{table}\vspace{-1em}

\section{Conclusion}
In this paper, we evaluated semantic and fixed-size chunking strategies in RAG systems across document retrieval, evidence retrieval, and answer generation. Semantic chunking occasionally improved performance, particularly on stitched datasets with high topic diversity. However, these benefits were highly context-dependent and did not consistently justify the additional computational cost. On non-synthetic datasets that better reflect real-world documents, fixed-size chunking often performed better. Overall, our results suggest that fixed-size chunking remains a more efficient and reliable choice for practical RAG applications. The impact of chunking strategy was often overshadowed by other factors, such as the quality of embeddings, especially when computational resources are limited or when working with standard document structures.

\clearpage
\newpage
\section*{Limitations}
\secname{Sentence-level Chunking}Our study focuses on sentence-level chunking, where documents are split into individual sentences, and each sentence is treated as a segment for grouping. This approach results in sentence embeddings that lack contextual information. While we attempted to address this by overlapping sentences in Fixed-size Chunker and incorporating positional distance in Semantic Chunker (global), the embeddings themselves remained context-free. Further exploration of contextual embeddings is necessary before definitively concluding the limitations of semantic chunking.\vspace{0.5em}

\secname{Lack of Chunk Quality Measures}As noted in Section \ref{sec:results}, while the output chunks differed between methods, retrieval and generation performances were similar across chunkers. In addition to the influence of embedding models, the absence of direct chunk quality metrics likely contributed to this issue. Having ground-truth query-chunk relevance scores would provide more accurate evaluations than relying solely on document or evidence mapping.\vspace{0.5em}

\secname{Lack of Suitable Datasets}Despite testing multiple datasets, our selection was constrained by a lack of comprehensive datasets. An ideal dataset would include long documents representative of real-world use cases, diverse query types, human-generated answers, query-document relevance scores, and human-labeled evidence sentences. Our synthetic documents had artificially high topic diversity due to random stitching, potentially leading to unreliable results. Additionally, the answer sets in RAGBench \cite{ragbench} were generated by LLMs, which may not accurately assess chunk quality. A dataset containing all these elements is needed for a more thorough evaluation of chunking strategies.

\bibliography{custom}

\clearpage
\newpage
\appendix

\begin{table*}[ht]
\centering
\begin{tabular}{l|c|c|r|r|r|r}
Dataset & Type & Split & $\#$D & S/D(*) & S/D & D/Q \\
\hline
Miracl \cite{miracl}               & Stitched & train & 1184 & 102 & 4   & 3 \\
NQ \cite{nq}                       & Stitched & test  & 488  & 88  & 5   & 1 \\
Scidocs \cite{scidocs}             & Stitched & test  & 1692 & 88  & 8   & 5 \\
Scifact \cite{scifact}             & Stitched & test  & 420  & 99  & 8   & 1 \\
BioASQ \cite{bioasq}               & Stitched & train & 2368 & 93  & 9   & 6 \\
NFCorpus \cite{nfcorpus}           & Stitched & test  & 364  & 118 & 12  & 37 \\
HotpotQA \cite{hotpotqa}           & Original & test  & 800  & 20  & 20  & 2 \\
MSMARCO \cite{msmarco}             & Original & dev   & 398  & 64  & 64  & 1 \\
ConditionalQA \cite{conditionalqa} & Original & dev   & 652  & 120 & 120 & 1 \\
Qasper \cite{qasper}               & Original & test  & 416  & 130 & 130 & 1 \\
\end{tabular}
\caption{Datasets for Document Retrieval. "$\#$D" means the number of selected long documents. "S/D" means the average number of sentences per document (before stitching). "S/D(*)" means the average number of sentences per long document (after stitching). "D/Q" means the average number of relevant long documents per query. The synthesized datasets are labeled as "Synthetic".}
\label{tab:datasets_doc_ret}
\vspace{1em}

\begin{tabular}{l|c|c|r|r|r}
Dataset & Split & $\#$D & S/D & E/Q \\
\hline
ExpertQA \cite{expertqa}           & test  & 777 & 20 & 12 \\
DelucionQA \cite{delucionqa}       & test  & 235 & 23 & 9 \\
TechQA \cite{techqa}               & test  & 648 & 49 & 15 \\
ConditionalQA \cite{conditionalqa} & dev   & 652 & 120 & 5 \\
Qasper \cite{qasper}               & test  & 416 & 130 & 4 \\
\end{tabular}
\caption{Datasets for Evidence Retrieval and Answer Generation. "$\#$D" means the number of selected long documents. "S/D" means the average number of sentences per long document. "E/Q" means the average number of evidence sentences per query.}
\label{tab:datasets_evi_ret}
\vspace{1em}

\centering
\begin{tabular}{l|r|r}
Name & Rank & Model Size (millions) \\
\hline
\texttt{dunzhang/stella$\_$en$\_$1.5B$\_$v5} \cite{stella} &   3 & 1543 \\
\texttt{BAAI/bge-large-en-v1.5} \cite{bge}                 &  36 & 335 \\
\texttt{all-mpnet-base-v2} \cite{mpnet}                    & 105 & 110 \\
\end{tabular}
\caption{Embedding models used in the experiments. "Rank" represents the rank of the model on the MTEB Leaderboard \cite{mteb}. "Model Size" represents the number of parameters in the embedding model.}
\label{tab:embedders}
\end{table*}

\section{Clustering methods for Clustering-based Semantic Chunker}\label{sec:clustering}
We applied single-linkage agglomerative clustering to sentence embeddings in two stages. First, we computed a distance matrix where each entry represents the distance between pairs of sentences in the document. Second, we iteratively formed clusters by merging sentence pairs with the smallest distances, ensuring that the resulting cluster does not exceed a predefined maximum chunk size. This process continued until all distances had been processed, after which we relabeled the merged clusters.

To address challenges encountered during experimentation, we implemented the following adjustments:\vspace{0.5em}

\secname{Chunk Size Constraint}Without a size constraint, this chunker tends to form one large chunk while leaving a few isolated sentences as individual chunks. To avoid this, we imposed a maximum chunk size threshold that directly depends on the number of chunks and the total number of sentences in the input document.\vspace{0.5em}

\secname{Distance Threshold for Stopping}To prevent isolated sentences from being grouped arbitrarily, we introduced a distance threshold. Once this threshold is exceeded, clustering stops, and any remaining sentences are left ungrouped. In this paper, the threshold was set to be 0.5. \vspace{0.5em}

A limitation of the single-linkage method is its requirement to specify the number of clusters, which can be difficult without prior knowledge. To mitigate this, we also experimented with DBSCAN \cite{dbscan}, a density-based clustering method that adjusts the number of clusters dynamically based on the density of sentence embeddings. DBSCAN follows the same initial steps as single-linkage clustering but replaces the merging process with density-based clustering.

\section{Hyperparameters}\label{sec:hyperparam}
\secname{Fixed-size Chunker} We tested two hyperparameters: the number of chunks and the number of overlapping sentences between consecutive chunks. For the number of chunks, we tested integer values between 2 and 10 to observe performance changes with different chunk sizes. For the overlapping sentences, we tested two settings: 0 or 1. If set to 1, one sentence overlaps between consecutive chunks; if set to 0, there is no overlap.

\secname{Breakpoint-based Semantic Chunker} We tested two hyperparameters: the type of breakpoint threshold and the threshold amount. Sentences were split into chunks when the distance between consecutive sentences exceeded a predefined threshold. We evaluated four relative threshold types from \cite{greg}:
\begin{itemize}
\item \textbf{Percentile}: The nth percentile of the linear interpolation of the distance array. We tested [10, 30, 50, 70, 90].
\item \textbf{Standard deviation}: The mean of the linear interpolation plus a fraction of the standard deviation. We tested [1, 1.5, 2, 2.5, 3].
\item \textbf{Interquartile}: The mean of the linear interpolation plus a fraction of the interquartile range. We tested [0.5, 0.75, 1, 1.25, 1.5].
\item \textbf{Gradient}: The nth percentile of the second-order accurate difference in the distance array. We tested [10, 30, 50, 70, 90].
\end{itemize}

Additionally, we tested two absolute versions of "Percentile" and "Gradient": \begin{itemize}
\item \textbf{Distance}: A cosine distance threshold value. We tested [0.1, 0.2, 0.3, 0.4, 0.5] based on empirical distance values.
\item \textbf{Gradient}: A threshold value based on the second-order accurate difference. We tested [0.01, 0.05, 0.1, 0.15, 0.2] based on empirical gradient values.
\end{itemize}

Note that the number of chunks or chunk size is not tunable in the Breakpoint-based Semantic Chunker. \vspace{0.5em}

\secname{Clustering-based Semantic Chunker} For the single-linkage chunker, we tested two hyperparameters: $\lambda$, which controls the weight of the positional distance in the overall distance calculation, and the number of chunks, as in the Fixed-size Chunker. We tested [0, 0.25, 0.5, 0.75, 1] for $\lambda$.

For the DBSCAN chunker, we evaluated three hyperparameters: $\lambda$, similar to single-linkage; EPS, the maximum distance between two samples for them to be considered part of the same neighborhood; and "min$\_$samples", the minimum number of samples required in a neighborhood for a point to be classified as a core point. For EPS, we tested [0.1, 0.2, 0.3, 0.4, 0.5]. For "min$\_$samples", we tested [1, 2, 3, 4, 5].

\section{Document Stitching and Dataset Choices} \label{sec:data_synthesis}
Most document retrieval datasets consist of short documents (fewer than 20 sentences), which are inadequate for effectively evaluating chunkers. Initially, we experimented with datasets from BEIR \cite{beir}, but the short length of these documents showed no performance improvement with chunking. Short documents lack the complexity required to assess how chunkers manage context and semantic coherence across longer spans of text.\vspace{0.5em}

To overcome this limitation, we created long documents by stitching shorter documents from existing datasets. Each stitched document contains approximately 100 sentences, better reflecting real-world long-document retrieval scenarios. In this setup, if a short document is relevant to a query, the corresponding stitched long document is considered relevant. This creates a coarser granularity for document retrieval and motivated the need for the evidence retrieval experiment, which offers a finer level of evaluation.\vspace{0.5em}

We selected datasets based on diversity in document topics and query types. Keyword-specific queries tend to favor lexical search, which can degrade the performance of semantic search methods. For the document retrieval task, we used the datasets listed in Table \ref{tab:datasets_doc_ret}, including NFCorpus, NQ, HotpotQA, Scidocs, and Scifacts from BEIR \cite{beir}.\vspace{0.5em}

For evidence retrieval and answer generation, we used the datasets listed in Table \ref{tab:datasets_evi_ret}. No stitched document was used.

\section{Choice of Evaluation Metrics}\label{sec:metrics}
\secname{Document Retrieval}Retrieval can be viewed as two tasks: classification and ranking. In this paper, a document is considered retrieved if any chunk from it is retrieved, irrespective of the query-chunk relevance score. This approach shifts the focus from query-chunk relevance to query-document evaluation, reducing the influence of ranking metrics such as NDCG, MAP, or MRR.

\begin{itemize}
\item \textbf{Recall@k}: Fraction of relevant documents retrieved within the top-k chunks, over all relevant documents.
\item \textbf{Precision@k}: Fraction of relevant documents retrieved within the top-k chunks, over all retrieved documents.
\item \textbf{F1@k}: The harmonic mean of precision and recall.
\end{itemize}

In typical retrieval experiments, recall is often the primary metric. However, our setup requires balancing recall with precision and F1 score. Since the number of retrieved chunks is fixed but the number of retrieved documents varies, precision and F1 are crucial. For instance, if five chunks are retrieved for a query with only one relevant document, retrieving all five chunks from this document would result in 100$\%$ recall and precision. However, if only one chunk is relevant and the rest are from irrelevant documents, the recall remains 100$\%$, but precision drops, leading to a different quality of retrieval. In such cases, the F1 score better captures this trade-off by balancing recall and precision.\vspace{1em}

\secname{Evidence Retrieval}In evidence retrieval, recall and precision are sensitive to chunk size when considered separately. Larger chunks tend to have higher recall, as they are more likely to contain evidence sentences, but also lower precision, as they may include more irrelevant sentences. Larger chunks are often less desirable as they introduce more noise. For example, "No Chunker" will consistently have the highest recall and lowest precision, as it treats entire documents as single chunks. The F1 score helps balance these biases, providing a better indicator of whether the chunker produces appropriately sized chunks that capture relevant evidence. Therefore, we focus on F1 scores in our analysis.

\begin{itemize}
\item \textbf{Recall@k}: Fraction of retrieved evidence sentences over all evidence sentences.
\item \textbf{Precision@k}: Fraction of retrieved evidence sentences over all retrieved sentences.
\item \textbf{F1@k}: The harmonic mean of precision and recall.
\end{itemize}

\secname{Answer Generation}Generated answers were assessed using BERTScore for semantic similarity between generated and actual answers, and cosine similarity between the queries and generated answers. 
\begin{itemize}
\item \textbf{BERTScore} \cite{bertscore}: A measure of the semantic similarity between generated answers and reference answers using contextual embeddings. We used the best model \texttt{microsoft/deberta-xlarge-mnli} for calculating this score.
\item \textbf{QA Similarity}: The cosine similarity between the query and generated answer, providing a measure of consistency and correctness in relation to the original query.
\end{itemize}

\clearpage
\newpage
\section{Additional Results and Analyses}\label{sec:additional_results}
We present full results and analyses that are not reported in Section \ref{sec:results} in this section. See Table \ref{tab:f1_doc} for F1 scores at all $k$ values for document retrieval. See Table \ref{tab:f1_stella} for F1 scores at all $k$ values for evidence retrieval.

\subsection{Results for Answer Generation}\label{sec:generation}
As shown in Table \ref{tab:qa_cos}, semantic chunkers performed slightly better than Fixed-size Chunker in terms of QA cosine similarity. However, the differences are minimal, making it difficult to draw any definitive conclusions from the results.
\begin{table}[h]
\centering
\centering
\resizebox{\columnwidth}{!}{\begin{tabular}{l|r|r|r}
Dataset & Fixed-size & Breakpoint & Clustering \\
\hline
ExpertQA      & 0.81 & \textbf{0.82} & 0.81 \\
DelucionQA    & 0.82 & 0.82 & \textbf{0.82} \\
TechQA        & \textbf{0.89} & 0.88 & 0.89 \\
ConditionalQA & 0.36 & \textbf{0.36} & 0.36 \\
Qasper        & 0.44 & 0.44 & \textbf{0.44} \\
\end{tabular}}
\caption{QA Cosine Similarity for Answer Generation.}
\label{tab:qa_cos}
\end{table}

\subsection{Impact of Embedding Models}\label{sec:embedders}
The choice of embedding model significantly affected retrieval performance (See Table \ref{tab:embedders} for tested models). In the Evidence Retrieval experiment, \texttt{BAAI/bge-large-en-v1.5} outperformed \texttt{all-mpnet-base-v2} by 1.06$\%$ on F1@1 and 1.32$\%$ on F1@10, both statistically significant at the 5$\%$ level. \texttt{dunzhang/stella$\_$en$\_$1.5B$\_$v5} showed an average improvement of 7.44$\%$ over \texttt{BAAI/bge-large-en-v1.5} across all F1 values. This result was statistically significant with $p=1.59\times10^{-5}$, highlighting the critical role of embedding models in retrieval tasks. See Tables \ref{tab:f1_stella}-\ref{tab:f1_mpnet} for full F1 scores from the three embedding models on evidence retrieval.

\subsection{Hyperparameter Analysis}
\begin{figure*}
\centering
\includegraphics[width=0.8\columnwidth]{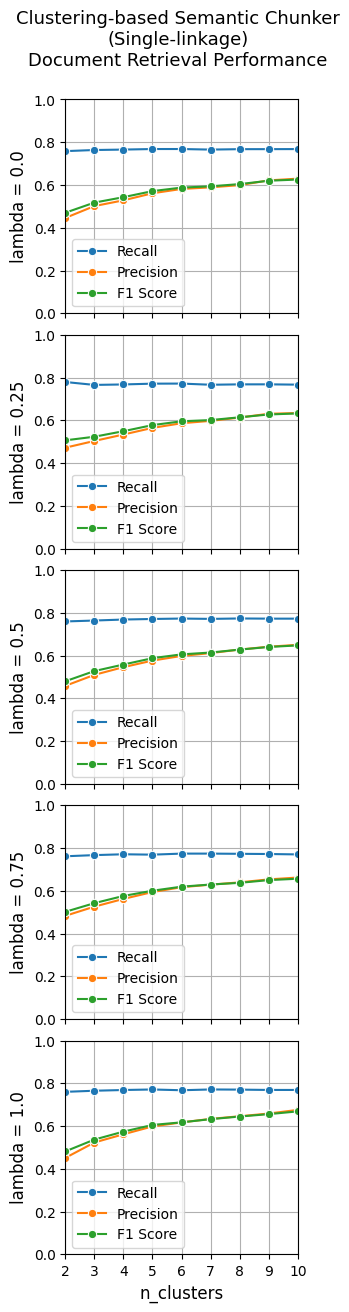}
\includegraphics[width=0.8\columnwidth]{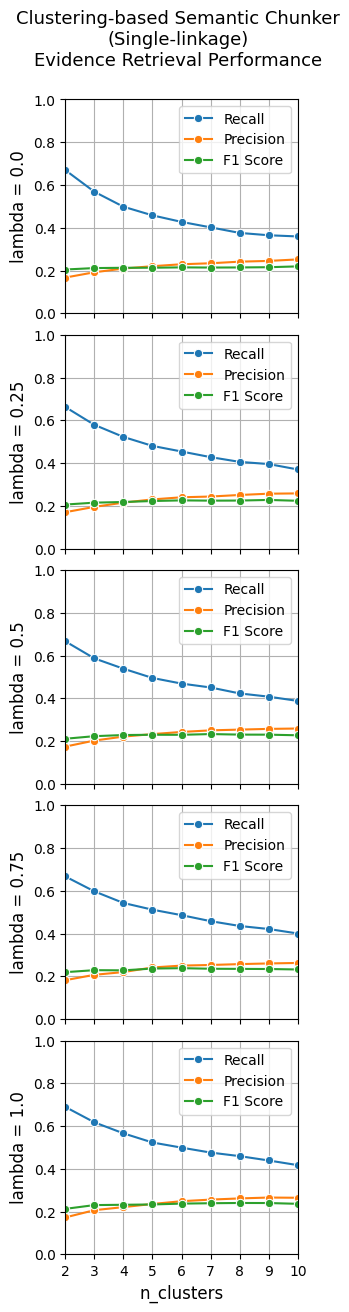}
\caption{Performance vs. hyperparameter values for Clustering-based Semantic Chunker (Single-linkage). Left: Document Retrieval; Right: Evidence Retrieval. The x-axis shows \texttt{n$\_$clusters}, and the y-axis shows the metric value. Each subplot’s y-label indicates the fixed hyperparameter value, with $\lambda$ increasing from top to bottom.}
\label{fig:hyperparam_single}
\end{figure*}

\begin{figure*}
\centering
\includegraphics[width=0.8\columnwidth]{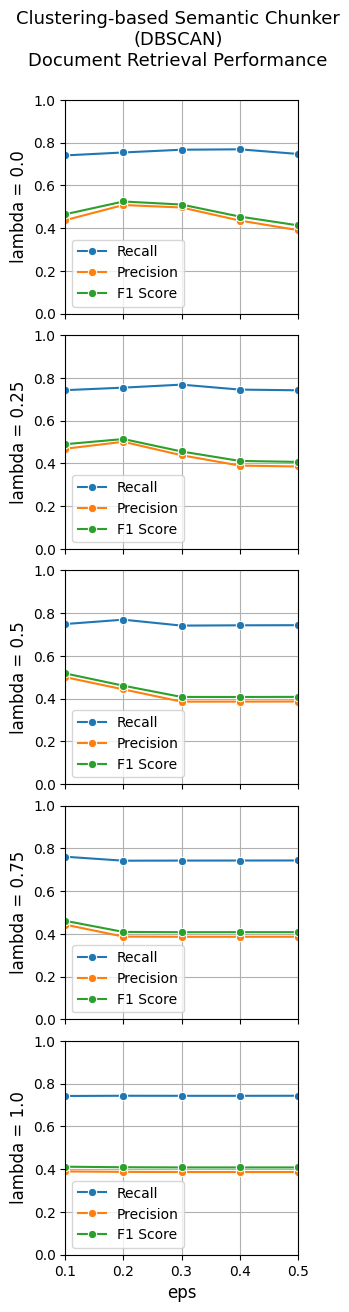}
\includegraphics[width=0.8\columnwidth]{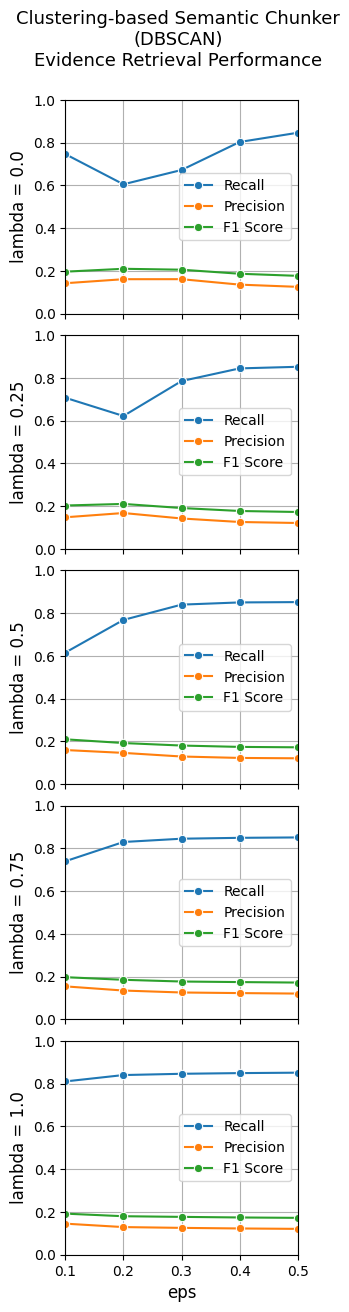}
\caption{Performance vs. hyperparameter values for Clustering-based Semantic Chunker (DBSCAN). Left: Document Retrieval; Right: Evidence Retrieval. The x-axis shows \texttt{eps}, and the y-axis shows the metric value. Each subplot’s y-label indicates the fixed hyperparameter value, with $\lambda$ increasing from top to bottom.}
\label{fig:hyperparam_dbscan}
\end{figure*}

\begin{figure*}
\centering
\includegraphics[width=0.7\columnwidth]{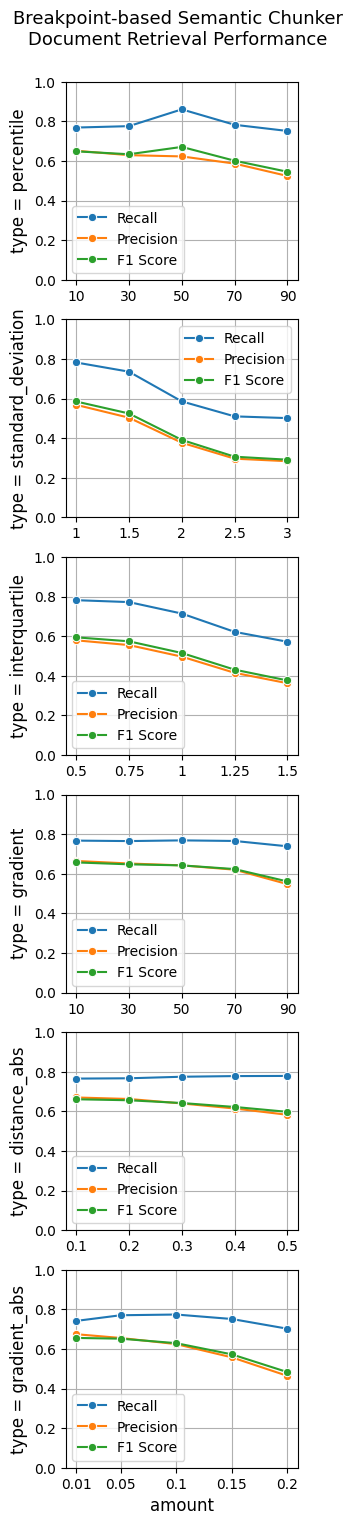}
\includegraphics[width=0.7\columnwidth]{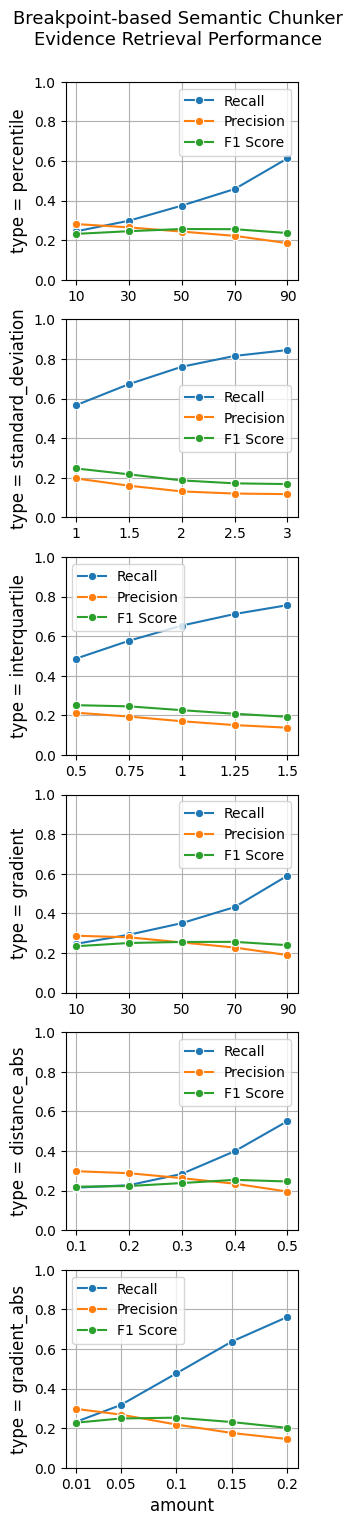}
\caption{Performance vs. hyperparameter values for Breakpoint-based Semantic Chunker. Left: Document Retrieval; Right: Evidence Retrieval. The x-axis shows \texttt{n$\_$clusters}, and the y-axis shows the metric value. Each subplot’s y-label indicates the breakpoint threshold type.}
\label{fig:hyperparam_cleavage}
\end{figure*}

\begin{figure*}
\centering
\includegraphics[width=\columnwidth]{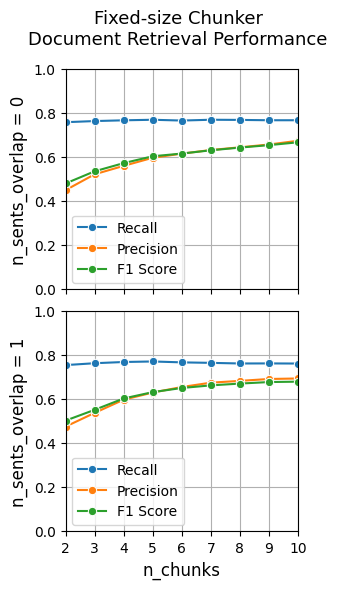}
\includegraphics[width=0.985\columnwidth]{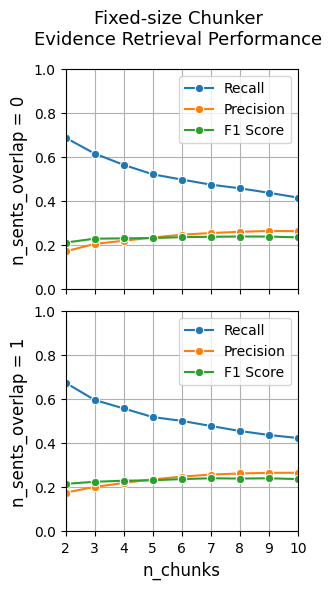}
\caption{Performance vs. hyperparameter values for Fixed-size Chunker. Left: Document Retrieval; Right: Evidence Retrieval. The x-axis shows \texttt{n$\_$chunks}, and the y-axis shows the metric value. Each subplot’s y-label indicates the fixed hyperparameter value, with \texttt{n\_sents\_overlap} increasing from top to bottom.}
\label{fig:hyperparam_fixedsize}
\end{figure*}

For Figures \ref{fig:hyperparam_single}-\ref{fig:hyperparam_fixedsize}, all scores are normalized and averaged across datasets and $k$ values. We aimed to identify chunker configurations that perform well across various datasets and $k$ values, making it logical to average the results. The title of each plot row indicates the chunker and experiment being analyzed, while each subplot title specifies the fixed hyperparameter. The y-axis shows the metric score, and the x-axis represents the hyperparameter being analyzed. Blue lines denote recall, orange lines represent precision, and green lines indicate the F1 score.\vspace{0.5em}

\secname{Clustering-based Semantic Chunker (Single-linkage)}As \texttt{n$\_$clusters} increases, the average chunk size decreases. This has little effect on document retrieval since chunks are mapped to their source documents regardless of size. However, Figure \ref{fig:hyperparam_single} shows that while recall remains steady, precision rises significantly as chunk size decreases, even when $\lambda=1$ (the Fixed-size Chunker case). This occurs due to a drop in the number of retrieved documents as smaller chunks from the same document are retrieved.

No clear trend for $\lambda$ was observed, indicating that shifting the weight between semantic and positional information does not significantly affect document retrieval. This suggests two possibilities: (1) Sentences close in position are often semantically similar; (2) Chunks with non-contiguous, yet semantically similar sentences do not enhance document retrieval.

In Figure \ref{fig:hyperparam_single}, evidence retrieval shows an inverse trend. As chunk size decreases, fewer sentences are retrieved, lowering the chance of retrieving evidence sentences and causing a sharp decline in recall. Thus, the F1 score remains relatively unchanged.

In addition, Figure \ref{fig:hyperparam_single} shows that as $\lambda$ approaches 1 (representing the Fixed-size Chunker), the F1 score (green line) gradually increases, indicating that positional information contributed more to retrieval performance than semantic similarity, likely because core evidence sentences were often located close together.\vspace{0.5em}

\secname{Clustering-based Semantic Chunker (DBSCAN)}As \texttt{EPS} increases, the threshold for grouping samples into the same cluster loosens, increasing average chunk size. As seen in Figure \ref{fig:hyperparam_dbscan}, this leads to a decrease in precision and an increase in recall for document and evidence retrieval, respectively, similar to the single-linkage case.\vspace{0.5em}

\secname{Breakpoint-based Semantic Chunker}As the distance threshold between consecutive sentences increases, fewer breakpoints appear, resulting in larger chunks. Regardless of the threshold type, it ultimately determines chunk size. In Figure \ref{fig:hyperparam_cleavage}, we observe similar trends to Figure \ref{fig:hyperparam_single} and \ref{fig:hyperparam_dbscan}: as chunk size increases, precision decreases in both retrieval tasks, while recall increases sharply for evidence retrieval. The rise in standard deviation is expected, as values from standard deviation-based thresholds are generally higher than those from percentiles or interquartile ranges.\vspace{0.5em}

\secname{Fixed-size Chunker}Figure \ref{fig:hyperparam_fixedsize} shows results for the Fixed-size Chunker. The trends mirror those seen in other chunkers. Adding one overlapping sentence between chunks does not notably improve performance, indicating that a single overlapping sentence is insufficient to significantly boost contextual coherence.

\subsection{Chunk Inspection}\label{sec:inspect}
\begin{figure*}
\centering
\includegraphics[width=\linewidth]{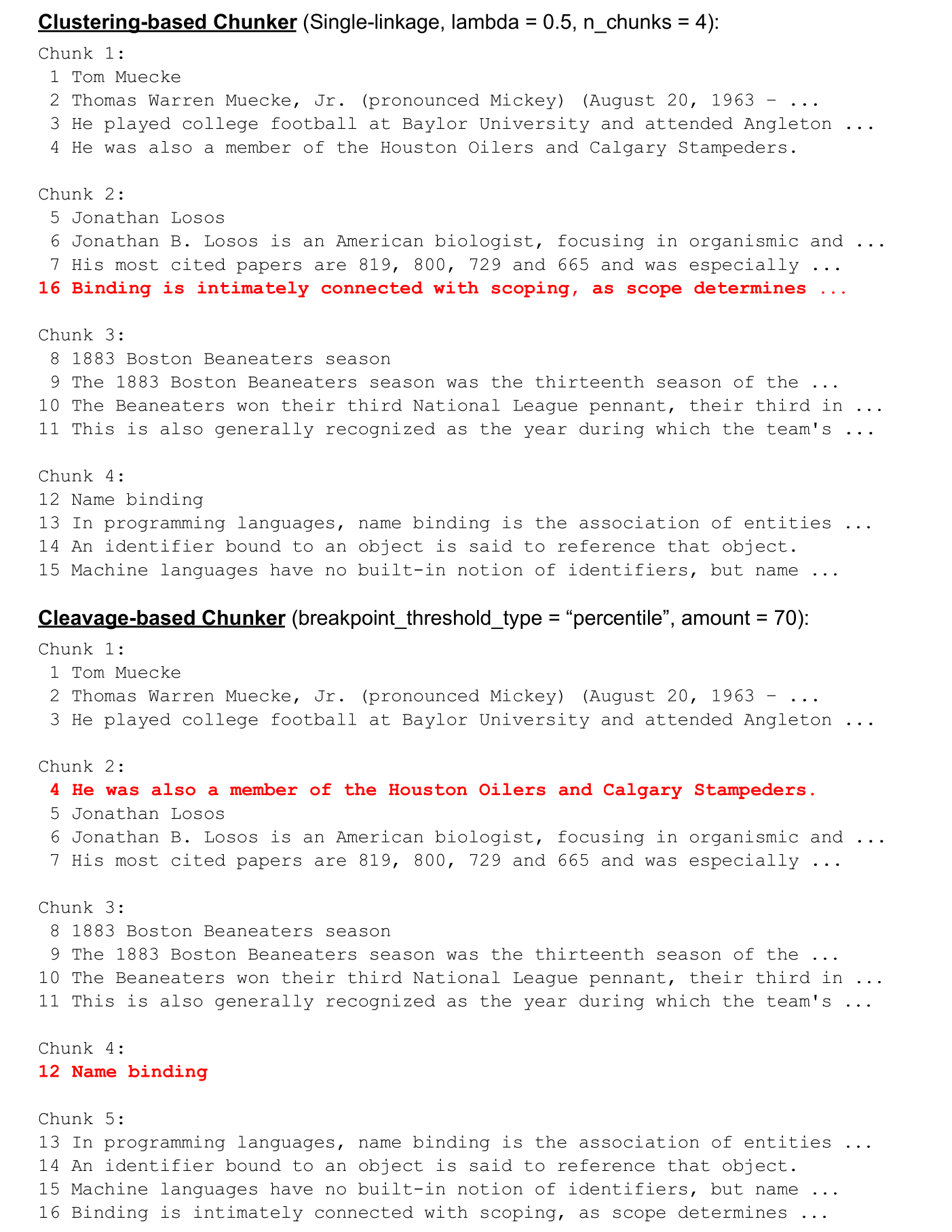}
\caption{Example of chunking a stitched document using different chunkers. Each line shows a sentence and its original index in the document. Bold red lines indicate errors where a sentence is incorrectly assigned to a chunk. The configuration listed next to each chunker name represents the optimal setup for minimizing errors.}
\label{fig:inspect_stitched}
\end{figure*}

\begin{figure*}
\centering
\includegraphics[width=\linewidth]{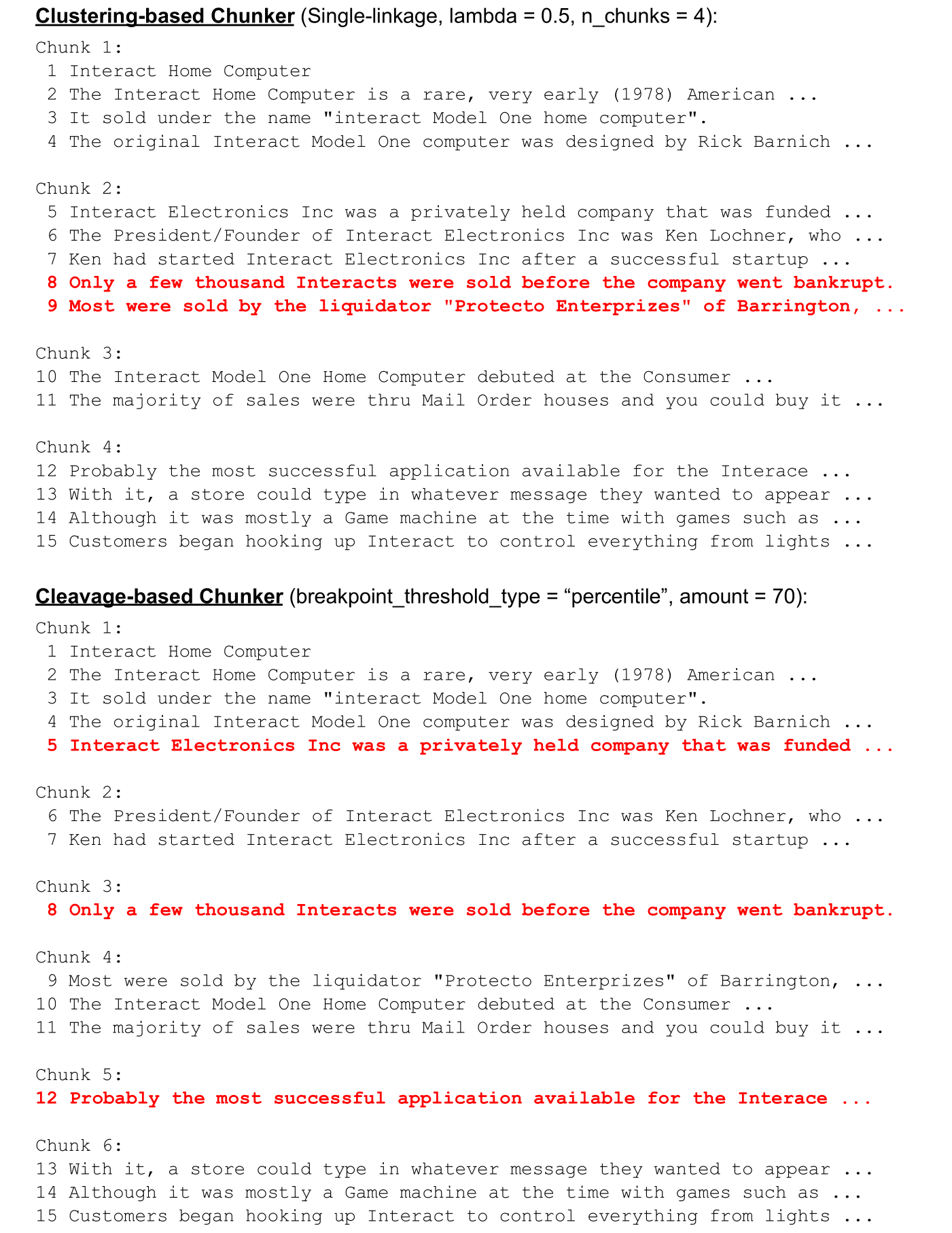}
\caption{Example of chunking a normal document using different chunkers. Each line shows a sentence and its original index in the document. Bold red lines indicate errors where a sentence is incorrectly assigned to a chunk. The configuration listed next to each chunker name represents the optimal setup for minimizing errors.}
\label{fig:inspect_normal}
\end{figure*}

We examined the output chunks to (1) confirm that different chunkers were functioning as intended, and (2) investigate the reasons behind performance differences. BEIR's HotpotQA dataset \cite{beir, hotpotqa} was selected for its reasonably sized documents. We randomly sampled five documents, stitching the first four together to form a stitched document (Figure \ref{fig:inspect_stitched}), and keeping the fifth as a normal document (Figure \ref{fig:inspect_normal}. The document IDs are:
\begin{itemize}
\item Stitched: 44547136, 14115210, 5580754, 54045118.
\item Normal: 30214079.
\end{itemize}

\secname{Inspection on Stitched Documents}In Figure \ref{fig:inspect_stitched}, Documents 1 and 3 have four sentences each, while Documents 2 and 4 contain three and five sentences, respectively. The Fixed-size Chunker, which ignores semantic relationships and document structure, frequently misassigned sentences, leading to errors that propagated through subsequent chunks. For instance, a sentence from Document 3 was appended to Document 2, illustrating the limitations of Fixed-size Chunking with stitched documents containing numerous short segments. This explains its poor performance under such conditions. However, simply splitting the document into structured sections before applying fixed-size chunking will solve this issue.

In contrast, both semantic chunkers performed better on stitched documents, but still had issues. The Clustering-based Chunker made one error by grouping Sentence 16 (the last sentence of Document 4) into Chunk 2. This happened because, despite the large positional distance, the semantic similarity was high, causing the sentence to be incorrectly included. Without considering positional structure like the Fixed-size and Breakpoint-based Chunkers, the Clustering-based Chunker often mixed sentences from different documents. While this might be useful for multi-document tasks \cite{multidoc, multidoc2}, it was problematic here, leading to worse performance when many short documents were stitched together.

The Breakpoint-based Chunker also made errors. It could, like the Fixed-size Chunker, group a sentence with a different chunk due to low semantic similarity with neighboring sentences, as seen with Sentence 4 being moved to Chunk 2. This shows the advantage of the joint distance measure in Equation \ref{eq:1}, which prevented this error for the Clustering-based Chunker. Moreover, controlling chunk size was challenging; higher thresholds led to overly large chunks, while lower thresholds resulted in single-sentence chunks lacking contextual information, such as Chunk 4's meaningless "Name binding" phrase.\vspace{0.5em}

\secname{Inspection on Normal Documents}In Figure \ref{fig:inspect_normal}, the document about "Interact Home Computer" was naturally divided into four sections, though this structure was not provided to the chunkers. The Fixed-size Chunker repeated its issue from stitched documents, occasionally grouping sentences from adjacent sections into the same chunk, and this error could be easily fixed by splitting the document by sections beforehand.

Although this example did not fully highlight the Clustering-based Chunker's limitations, it still demonstrated the downsides of relying solely on semantic similarity. Sentences 8-9, though belonging to Chunk 3, were grouped into Chunk 2 due to high semantic similarity. This showed that even with added positional information, semantic-based chunking could misgroup content that shared context, as these sentences were clearly about the sales of Interact Home Computer.

For the Breakpoint-based Chunker, errors seen in stitched documents were even more pronounced. Despite using the optimal configuration for each chunker (minimizing errors), Breakpoint-based Chunker still produced chunks containing only a single sentence, such as Chunk 3 and 5. Additionally, separating Sentences 5 and 6, which both discussed "Interact Electronics Inc," was an especially poor decision. These examples underscore that semantic similarity alone is not a reliable measure for effective chunking, and it may be less useful than straightforward positional information.

\begin{sidewaystable*}[hp!]
\centering
\resizebox{\textheight}{!}{\begin{tabular}{l|ccc|ccc|ccc|ccc}
\hline
Metric & \multicolumn{3}{|c|}{F1@1} & \multicolumn{3}{|c|}{F1@3} & \multicolumn{3}{|c|}{F1@5} & \multicolumn{3}{|c}{F1@10} \\
\hline
Chunker & Fixed-size & Breakpoint & Clustering & Fixed-size & Breakpoint & Clustering & Fixed-size & Breakpoint & Clustering & Fixed-size & Breakpoint & Clustering \\
\hline
Miracl*       & 67.55 & \textbf{69.73} & 68.61 & 76.03 & \textbf{83.55} & 75.24 & 69.45 & \textbf{81.89} & 67.35 & 49.89 & \textbf{67.83} & 46.59 \\
NQ*           & \textbf{92.92} & 92.36 & 88.63 & 62.29 & \textbf{83.37} & 60.29 & 43.79 & \textbf{63.93} & 41.01 & 24.02 & \textbf{36.18} & 22.56 \\
Scidocs*      &  7.60 &  7.73 & \textbf{10.40} & 15.16 & 14.92 & \textbf{18.93} & 16.82 & 16.60 & \textbf{19.87} & 16.96 & 16.88 & \textbf{19.94} \\
Scifact*      & 55.07 & 53.38 & \textbf{55.09} & 43.97 & \textbf{52.91} & 46.60 & 35.27 & \textbf{36.27} & 35.70 & 22.33 & \textbf{27.59} & 22.32 \\
BioASQ*       & 53.09 & \textbf{55.95} & 53.14 & 61.92 & \textbf{70.74} & 61.84 & 61.86 & 61.87 & \textbf{62.49} & 54.37 & \textbf{56.82} & 55.44 \\
NFCorpus*     & 11.41 & \textbf{12.49} & 11.42 & 19.00 & 19.10 & \textbf{20.24} & 21.36 & 21.07 & \textbf{22.12} & 22.95 & 23.48 & \textbf{24.09} \\
HotpotQA      & 66.00 & 66.00 & \textbf{66.67} & 92.06 & 91.83 & \textbf{92.33} & \textbf{90.59} & 87.37 & 84.79 & \textbf{61.34} & 52.22 & 51.30 \\
MSMARCO       & \textbf{99.00} & 97.00 & 98.00 & \textbf{95.35} & 94.92 & 94.73 & \textbf{93.58} & 92.23 & 93.18 & \textbf{85.75} & 84.34 & 77.57 \\
ConditionalQA & \textbf{83.03} & 79.70 & 79.34 & \textbf{78.67} & 74.63 & 76.09 & \textbf{68.11} & 64.44 & 65.94 & \textbf{44.66} & 40.37 & 39.35 \\
Qasper        & \textbf{96.50} & 93.53 & 95.96 & \textbf{95.21} & 92.20 & 95.14 & \textbf{90.99} & 89.27 & 90.77 & 68.86 & \textbf{69.59} & 62.41 \\
\end{tabular}}
\caption{F1 scores for all $k$ values for Document Retrieval ($\%$). Datasets marked with * are stitched. Rows are sorted by the average number of sentences per document (before stitching) in ascending order for easier comparison.}
\label{tab:f1_doc}
\vspace{5em}
\end{sidewaystable*}

\begin{sidewaystable*}[hp!]
\centering
\resizebox{\textheight}{!}{\begin{tabular}{l|ccc|ccc|ccc|ccc}
\hline
Metric & \multicolumn{3}{|c|}{F1@1} & \multicolumn{3}{|c|}{F1@3} & \multicolumn{3}{|c|}{F1@5} & \multicolumn{3}{|c}{F1@10} \\
\hline
Chunker & Fixed-size & Breakpoint & Clustering & Fixed-size & Breakpoint & Clustering & Fixed-size & Breakpoint & Clustering & Fixed-size & Breakpoint & Clustering \\
\hline
ExpertQA      & \textbf{42.43} & 35.25 & 40.83 & 48.67 & 48.49 & \textbf{48.73} & \textbf{47.11} & 47.08 & 46.87 & 33.18 & \textbf{36.53} & 33.92 \\
DelucionQA    & \textbf{39.40} & 28.12 & 34.60 & 44.18 & \textbf{45.43} & 44.05 & 43.05 & 43.24 & \textbf{43.36} & \textbf{37.29} & 36.16 & 36.32 \\
TechQA        & \textbf{39.38} & 29.27 & 31.68 & \textbf{28.98} & 28.49 & 27.96 & \textbf{28.98} & 28.49 & 27.96 & \textbf{16.92} & 16.76 & 14.51 \\
ConditionalQA & 23.14 & \textbf{23.61} & 22.15 & 19.81 & \textbf{22.01} & 17.32 & 18.23 & \textbf{19.83} & 19.14 & 14.56 & \textbf{15.41} & 15.25 \\
Qasper        & 8.22 & \textbf{8.58} & 8.36 & \textbf{9.67} & 8.83 & 8.75 & \textbf{8.66} & 8.16 & 8.50 & \textbf{6.99} & 6.78 & 6.52 \\
\end{tabular}}
\caption{F1 scores for all $k$ values for Evidence Retrieval ($\%$), from \texttt{dunzhang/stella$\_$en$\_$1.5B$\_$v5}.}
\label{tab:f1_stella}
\vspace{3em}

\resizebox{\textheight}{!}{\begin{tabular}{l|ccc|ccc|ccc|ccc}
\hline
Metric & \multicolumn{3}{|c|}{F1@1} & \multicolumn{3}{|c|}{F1@3} & \multicolumn{3}{|c|}{F1@5} & \multicolumn{3}{|c}{F1@10} \\
\hline
Chunker & Fixed-size & Breakpoint & Clustering & Fixed-size & Breakpoint & Clustering & Fixed-size & Breakpoint & Clustering & Fixed-size & Breakpoint & Clustering \\
\hline
ExpertQA      & \textbf{40.34} & 33.39 & 38.31 & \textbf{44.60} & 43.71 & 44.36 & 42.25 & 41.46 & \textbf{43.03} & 29.54 & \textbf{31.89} & 29.28 \\
DelucionQA    & \textbf{33.88} & 27.02 & 32.73 & 42.10 & \textbf{43.58} & 40.79 & 40.85 & 40.89 & \textbf{41.22} & \textbf{37.29} & 36.16 & 35.99 \\
TechQA        & \textbf{34.90} & 28.25 & 29.57 & 23.09 & \textbf{23.92} & 22.24 & 19.82 & \textbf{20.90} & 19.27 & 13.00 & \textbf{13.25} & 12.92 \\
ConditionalQA & \textbf{20.09} & \textbf{20.09} & 19.40 & \textbf{18.24} & 16.93 & 14.27 & \textbf{14.83} & 13.89 & 10.73 & \textbf{10.72} & 9.24 & 6.39 \\
Qasper        & \textbf{7.80} & 6.20 & 5.34 & \textbf{6.88} & 6.83 & 6.76 & \textbf{6.59} & 6.43 & 5.70 & \textbf{4.99} & 4.71 & 4.34 \\
\end{tabular}}
\caption{F1 scores for all $k$ values for Evidence Retrieval ($\%$), from \texttt{BAAI/bge-large-en-v1.5}.}
\label{tab:f1_bge}
\vspace{3em}

\resizebox{\textheight}{!}{\begin{tabular}{l|ccc|ccc|ccc|ccc}
\hline
Metric & \multicolumn{3}{|c|}{F1@1} & \multicolumn{3}{|c|}{F1@3} & \multicolumn{3}{|c|}{F1@5} & \multicolumn{3}{|c}{F1@10} \\
\hline
Chunker & Fixed-size & Breakpoint & Clustering & Fixed-size & Breakpoint & Clustering & Fixed-size & Breakpoint & Clustering & Fixed-size & Breakpoint & Clustering \\
\hline
ExpertQA      & \textbf{40.96} & 32.75 & 38.83 & 42.76 & \textbf{43.39} & 43.38 & 41.78 & 41.56 & \textbf{42.07} & \textbf{31.82} & 29.64 & 31.37 \\
DelucionQA    & \textbf{38.02} & 32.22 & 31.67 & 39.78 & \textbf{42.22} & 39.68 & \textbf{41.31} & 35.34 & 41.04 & 35.94 & 27.77 & \textbf{36.11} \\
TechQA        & \textbf{31.04} & 23.30 & 27.24 & \textbf{24.62} & 23.41 & 24.60 & 19.42 & \textbf{21.24} & 19.56 & \textbf{16.56} & 14.07 & 12.21 \\
ConditionalQA & 18.01 & \textbf{20.87} & 17.73 & 14.73 & \textbf{18.65} & 14.24 & 11.67 & \textbf{16.09} & 11.05 & 7.25 & \textbf{12.95} & 7.11 \\
Qasper        & \textbf{8.09} & 6.92 & 6.98 & \textbf{6.97} & 6.23 & 6.67 & 6.56 & 5.98 & 6.24 & \textbf{4.23} & 4.12 & 3.62 \\
\end{tabular}}
\caption{F1 scores for all $k$ values for Evidence Retrieval ($\%$), from \texttt{all-mpnet-base-v2}.}
\label{tab:f1_mpnet}
\end{sidewaystable*}

\begin{figure*}
\centering
\includegraphics[width=\linewidth]{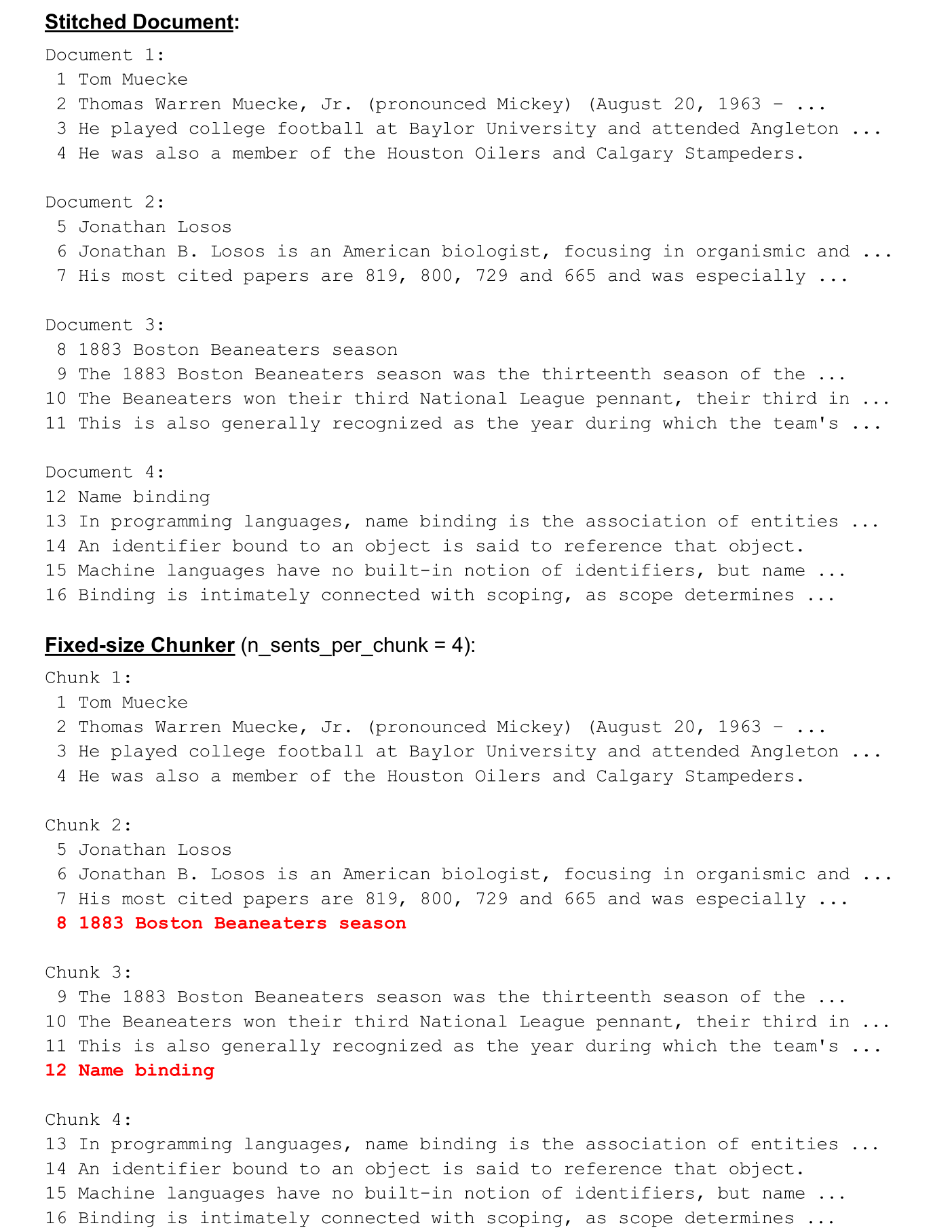}
\end{figure*}

\begin{figure*}
\centering
\includegraphics[width=\linewidth]{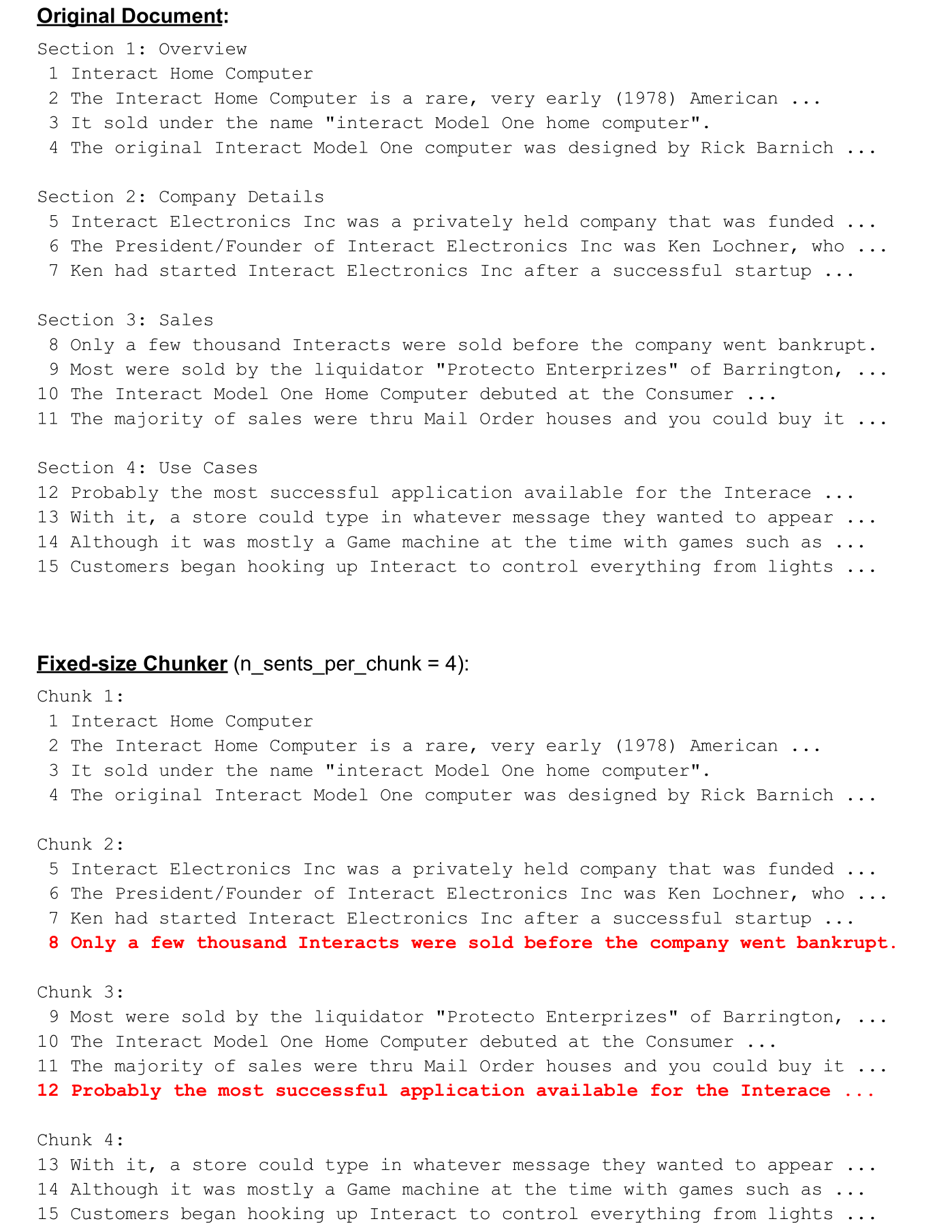}
\end{figure*}

\end{document}